\documentclass[10pt,twocolumn,letterpaper]{article}

\usepackage{wacv}
\usepackage{times}
\usepackage{epsfig}
\usepackage{graphicx}
\usepackage{amsmath}
\usepackage{amssymb}
\usepackage{xcolor}

\usepackage{booktabs}
\newcommand{\xdownarrow}[1]{%
  {\left\downarrow\vbox to #1{}\right.\kern-\nulldelimiterspace}
}

\usepackage{caption}
\wacvfinalcopy 

\ifwacvfinal
\usepackage[breaklinks=true,bookmarks=false]{hyperref}
\else
\usepackage[pagebackref=true,breaklinks=true,colorlinks,bookmarks=false]{hyperref}
\fi

\begin{document}

\title{Visual Speech-Aware Perceptual 3D Facial Expression Reconstruction from Videos}

\author{Panagiotis P. Filntisis\textsuperscript{1} \thinspace\quad George Retsinas \textsuperscript{1} \thinspace\quad Foivos Paraperas-Papantoniou\textsuperscript{4} \thinspace\quad Athanasios Katsamanis\textsuperscript{5} \\ \thinspace\quad Anastasios Roussos\textsuperscript{2,3} \thinspace\quad Petros Maragos\textsuperscript{1}  \\
\\
\textsuperscript{1}School of Electrical \& Computer Engineering, National Technical University of Athens, Greece\\
\textsuperscript{2}Institute of Computer Science (ICS), Foundation for Research \& Technology - Hellas (FORTH), Greece
\\
\textsuperscript{3}College of Engineering, Mathematics and Physical Sciences, University of Exeter, UK
\\
\textsuperscript{4}Imperial College London, UK 
\\
\textsuperscript{5}Institute for Language and Speech Processing, Athena R.C., Greece
}

\twocolumn[{%
\renewcommand\twocolumn[1][]{#1}%
\maketitle
\begin{center}
\vspace{-0.3cm}
    \centering
    \captionsetup{type=figure}
    \includegraphics[trim=0 0 0 0, clip, width=1\textwidth]{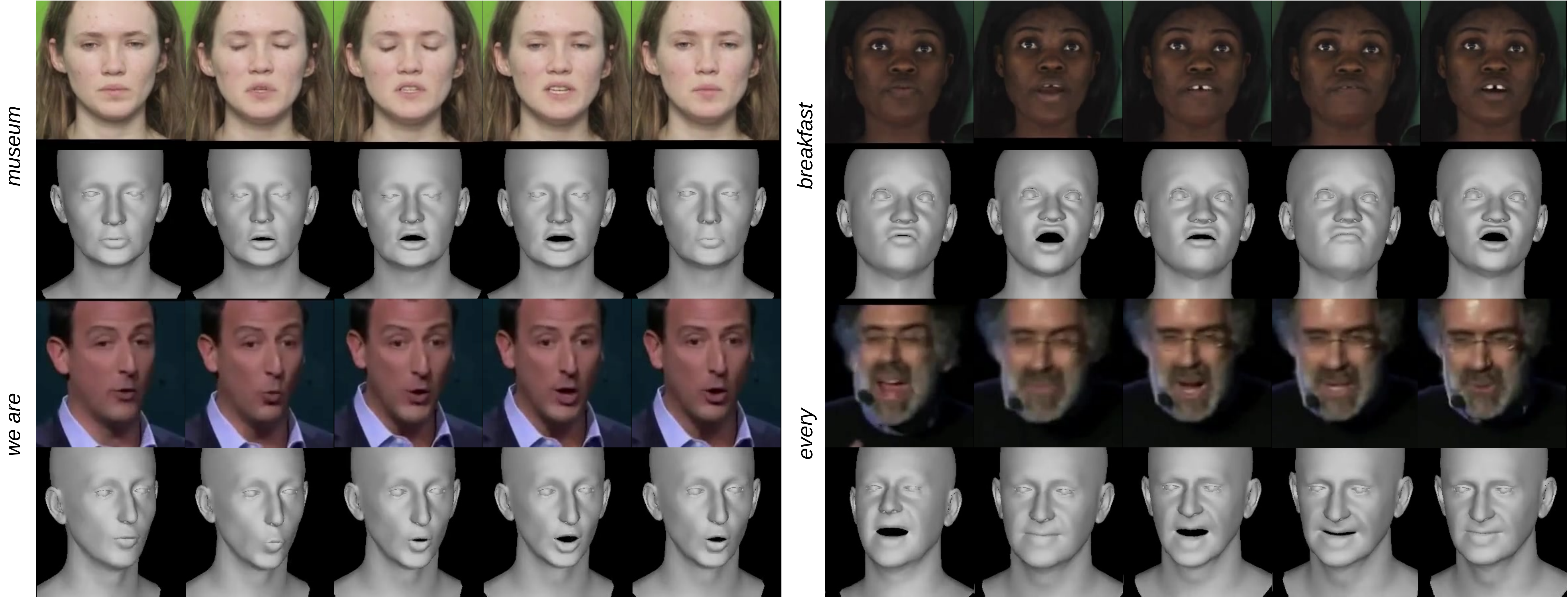}
    \vspace{-0.3cm}
    \captionof{figure}{Our method performs visual-speech aware 3D reconstruction so that speech perception from the original footage is preserved in the reconstructed talking head. On the left we include the word/phrase being said for each example. Please zoom-in for details and refer to \url{https://filby89.github.io/spectre} where you will also find many video examples with sound, code, and pretrained models.}
    \label{fig:ned_teaser}
\end{center}%
}]




\begin{abstract}
The recent state of the art on monocular 3D face reconstruction from image data has made some impressive advancements, thanks to the advent of Deep Learning. However, it has mostly focused on input coming from a single RGB image, overlooking the following important factors: 
a) Nowadays, the vast majority of facial image data of interest do not originate from single images but rather from videos, which contain rich dynamic information. b) Furthermore, these videos typically capture individuals in some form of verbal communication (public talks, teleconferences, audiovisual human-computer interactions, interviews, monologues/dialogues in movies, etc). 
When existing 3D face reconstruction methods are applied in such videos, the artifacts in the reconstruction of the shape and motion of the mouth area are often severe, since they do not match well with the speech audio. 

To overcome the aforementioned limitations, we present the first method for visual speech-aware perceptual reconstruction of 3D mouth expressions. We do this by proposing a ``lipread" loss, which guides the fitting process so that the elicited perception from the 3D reconstructed talking head resembles that of the original video footage. We demonstrate that, interestingly, the lipread loss is better suited for 3D reconstruction of mouth movements compared to traditional landmark losses, and even direct 3D supervision.  Furthermore, the devised method does not rely on any text transcriptions or corresponding audio, rendering it ideal for training in unlabeled datasets. We verify the efficiency of our method through exhaustive objective evaluations on three large-scale datasets, as well as subjective evaluation with two web-based user studies.
\end{abstract}

\section{Introduction}

During the last years, Deep Learning frameworks have succeeded in significantly increasing the accuracy of monocular 3D face reconstruction, even in cases of unconstrained image data. 
The current state of the art is able to robustly reconstruct fine details of the 3D facial geometry as well as yield a reliable estimation of the captured subject's facial anatomy. 
This is beneficial for a plethora of applications, such as  
augmented reality, performance capture, visual effects, photo-realistic video synthesis, human-computer interaction and personalized avatars, to name but a few. 

On the other hand, the vast majority of existing methods focus on 3D face reconstruction from a single RGB image, without exploiting the rich dynamic information that is inherent in humans' faces, especially during speech. 
But even the few methods that include some sort of dynamics modelling to reconstruct facial videos, do not explicitly model the strong correlation between mouth motions and articulated speech. 
At the same time, most facial videos of interest capture individuals involved in some form of verbal communication.  
When existing 3D face reconstruction methods are applied in this kind of videos, the artifacts in the reconstruction of the shape and motion of the mouth area are often severe and overwhelming in terms of human perception, the perceptual movements of the mouth that correspond to speech are not captured well. 

Arguably, a crucial factor for the limitations of existing methods is the fact that most methods use weak 2D supervision from landmarks predicted by face alignment methods as a form of guidance, 
e.g.~\cite{saito2016real,face2face,thies2016facevr, jackson2017large,booth20183d,tewari17MoFA,feng2021learning,yang2020facescape}. While these landmarks can yield a coarse estimation of the  facial shape, they fail to provide an accurate representation of the expressive details of a highly-deformable mouth region. 
It is also important to note that the shapes of the human mouth are perceptually correlated with speech and the realism of a 3D talking head is tightly coupled with the uttered sentence. As a result, a 3D model that talks without the lips closing when uttering the bi-labial consonants (i.e., /m/, /p/, and /b/), or with no lip-roundness when speaking a rounded-vowel (such as /o/ /u/) has a poor perceived naturalness. In EMOCA~\cite{danvevcek2022emoca}, some significant steps were done in improving the expressivity of the 3D reconstructed head, however the perceptual emotional consistency loss only affected those movements that correspond to emotions. Furthermore, this method did not predict the jaw parameters as well, resulting in poor articulation.

We conclude that, although speech perception from reconstructed 3D faces is important for various applications (e.g.~augmented and virtual reality, gaming, affective avatars etc.) \cite{hofer2020role,marin2020emotion,stuarteffect}, it is a commonly overlooked parameter in the existing literature. 
It is worth mentioning that the primary evaluation metric used by most existing methods is the distance of the predicted vertices of the model from the ground truth. However, geometric errors of facial/mouth expressions do not necessarily correlate with human perception~\cite{danvevcek2022emoca,mori2012uncanny,garrido2016corrective}. 


To overcome the limitations of the existing literature, this work tackles the problem of monocular 3D face reconstruction from a video, with a strong focus on the mouth area and its expressions and movements that are connected with speech articulation. 
 We highlight and address the fact that an accurate 3D reconstruction of a human talking in a video should retain those mouth expressions and movements that humans perceive to correspond to the speech. Our method leverages a SoTA model of lip reading in order to minimize the ``speech perceptual" distance between the rendered and the original input video. Our main contributions can be summarized as follows:
\\
$\bullet$ We design and implement the first (to the best of our knowledge) method for perceptual 3D reconstruction of human faces focusing on speech \textbf{without the need for text transcriptions of the corresponding audio}.
\\
$\bullet$ We devise a ``lipread" loss, which guides the fitting process so that the reconstructed face and especially the mouth area elicits similar perception to the viewer and feels more realistic when coupled with the corresponding audio.
\\
$\bullet$ We conduct extensive objective and subjective (user studies) evaluation that proves the significant increase in perception of the reconstructed talking head.  We also propose the usage of various lip-read metrics as an objective evaluation of the perception of human speech in reconstructed 3D heads.
\\
$\bullet$ We make the source code and models of our method publicly available.






\section{Related Work}
\paragraph{3D Models}
There is extensive literature in the fields of computer vision and graphics for creating and reconstructing 3D face models from various input sources (RGB, Depth) \cite{zollhoefer2018facestar, 3D_survey_past}. 
3D Morphable Models are by far the most widely-used choice, since they offer compact representations as well as a convenient decoupling of expression and identity variation, allowing better manipulation. The traditional 3DMMs were linear, PCA-based models of 3D shape variation, but several non-linear and deep learning-based extensions have been proposed during the last years \cite{tran2018nonlinear,bagautdinov2018modeling,abrevaya2019generative,cheng2019meshgan}. 
Some of the most popular 3D face models are the Basel Face Model~\cite{paysan20093d,8373814}, the FaceWarehouse model \cite{cao2013facewarehouse},
the FLAME~\cite{FLAME:SiggraphAsia2017}, and more recently the FaceScape~\cite{yang2020facescape}, and FaceVerse~\cite{wang2022faceverse} models. Usually, these models are built from large datasets of 3D scans of human faces.


\paragraph{Monocular 3D Face Reconstruction}
A common application of 3DMMs includes estimation of the model parameters that best fit to an RGB image. This can happen as a direct optimization procedure in an analysis-by-synthesis framework, e.g.~\cite{Blanz,aldrian2012inverse,Thies_2015,face2face,booth20183d,grassal2021neural}. However this is a computationally expensive procedure to run on novel images every time. For example, the recent FaceVerse method \cite{wang2022faceverse} needs $\approx 10$ minutes for detailed refinement. Due to this reason, multiple methods have emerged that formulate the problem as a regression from image data, leveraging the power of deep learning \cite{tewari17MoFA,jourabloo2016large,jackson2017large,tran2018extreme,gecer2019ganfit}. Combined with a reliable facial landmarker, this can lead to accurate results, even without the need for 3D supervision. 

For example, RingNet~\cite{RingNet:CVPR:2019}, performed 3D reconstruction using the FLAME model, by enforcing a shape-consistency loss between images of the shape subject, in order to decouple identity and expression. DECA~\cite{feng2021learning} further built upon RingNet and predicted parameters of the FLAME model jointly from a CNN, using multiple loss coefficients that tackle the lack of 3D ground truth. EMOCA~\cite{danvevcek2022emoca}  focused on the expressiveness of the reconstructed models, by adding an emotional perceptual loss and training a specific CNN that predicts the expression parameters of the 3DMM on a large emotional dataset (AffectNet). ExpNet~\cite{chang2018expnet} on the other hand generated pseudo-3DMM parameters by solving the optimization problem given an accurate 3D reconstruction of an image with a SoTA method, and then training a CNN to predict them, without the need for landmarks.
In 3DDFA~\cite{3ddfa_cleardusk, guo2020towards, zhu2017face}, face alignment and 3D reconstruction takes place concurrently, using Cascaded CNNs (expand more here). The recent MICA method \cite{zielonka2022mica} focused on accurate prediction of the identity parameters of a 3DMM, by employing a medium-scale 3D annotated dataset in conjunction with a large-scale 2D raw image dataset. Finally, DAD-3DHeads~\cite{dad3dheads}, provided one of the first large-scale 3D head datasets, that can be used for direct supervision of 3D reconstruction.

Even though the vast majority of methods reconstruct single face images or work on a frame-by-frame fashion on videos, there are a few methods that exploit the dynamic information of monocular face videos to constrain the subject's facial shape or impose temporal coherence on the face reconstruction \cite{cao2015real,garrido2016reconstruction,huber2016real,koujan2018combining,booth2018large}.

Our work is mostly similar to EMOCA~\cite{danvevcek2022emoca}, in the concept that it is concerned with both are concerned with perceptual reconstruction. In comparison however, EMOCA focuses on retaining affective information from images while our work focuses on accurate reconstruction of mouth and lips formation that correspond to speech production. Furthermore, EMOCA failed to accurately predict the jaw pose parameters which include opening and rotation of the mouth due to difficulties in convergence and kept the jaw pose fixed.


\paragraph{Mouth/Lip Reconstruction}
Some of the earliest works focusing on the dynamics of mouth and lips for 3D reconstruction include the works of Basu et al.~\cite{basu19983d,basu19983d1} which used a combined-statistical model, Gregor et al. \cite{10.1117/12.410873} who used markers to follow the lip motions, and Cheng et al.~\cite{cheng2010real} who performed mouth tracking from 2D mages using Adaboost and a Kalman filter. The most recent work concerned with lip tracking from video is the work of Garrido et al.~\cite{garrido2016corrective}, who achieved remarkable results of 3D lips reconstructed, using a high quality 3D stereo database for lip tracking and using the ground truth shapes along with radial basis functions to fix the results of reconstruction in 2D images.

\section{Method}
\subsection{Preliminaries}
Our work is based on the state-of-the-art DECA~\cite{feng2021learning} framework for monocular 3D reconstruction from static RGB images. As such we adopt the notation from the DECA paper. In the original DECA, given an input image $I$ a coarse encoder (a ResNet50 CNN) jointly predicts the identity parameters $\boldsymbol{\beta} \in \mathbb{R}^{100}$, neck pose and jaw $\boldsymbol{\theta} \in \mathbb{R}^{6}$, expression parameters $\boldsymbol{\psi} \in \mathbb{R}^{50}$, albedo $\boldsymbol{\alpha} \in \mathbb{R}^{50}$, lighting $\boldsymbol{I} \in \mathbb{R}^{27}$, and camera (scale and translation) $\boldsymbol{c} \in \mathbb{R}^{3}$. Note that these parameters are a subset of the parameters of the FLAME 3D face model. Afterwards, these parameters are used to render the predicted 3D face. DECA also included a detail encoder which predicted a latent vector associated with a UV-displacement map, that models high-frequency person-specific details such as wrinkles. More recently, EMOCA~\cite{danvevcek2022emoca} further built upon DECA by adding an extra expression encoder (ResNet50) which was used in order to predict the expression vector $\boldsymbol{\psi}$, so that the perceived emotion of the reconstructed face is similar to that of the original image. We use these two works as starting points and focus on designing an architecture that increases the perceived expressions of the input video, concentrating on the mouth area, leading to realistic articulation movements.

\subsection{Architecture}
\begin{figure*}[t]
\centering
\includegraphics[width=\textwidth]{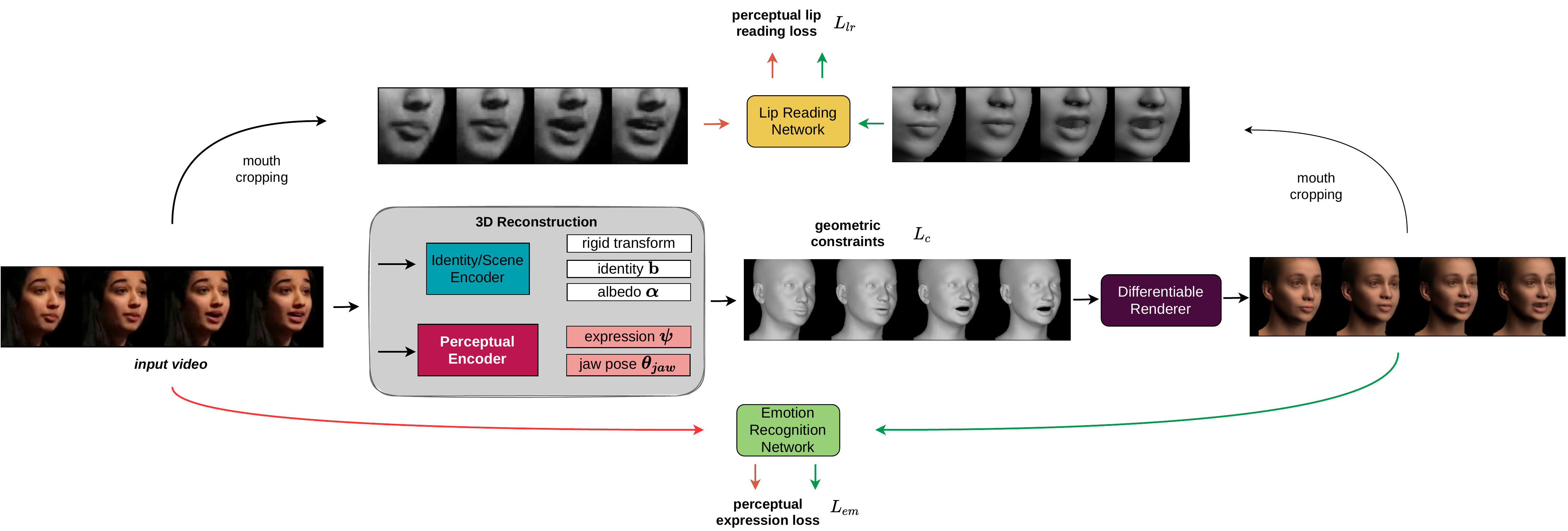}

\caption{Overview of our architecture for perceptual 3D reconstruction. The input video is first fed into the 3D reconstruction component, where a fixed encoder detects the scene parameters (camera, lighting), identity parameters (albedo/identity) and an initial estimate of the jaw and expression parameters. Then, a Mouth/Expression encoder predicts the refined facial expression parameters and jaw pose, and a differentiable renderer renders the predicted 3D shape. Finally, the mouth area is differentiably cropped in both the input and rendered image sequences and a lip reader is applied on both in order to estimate the perceptual lip reading loss between them. The same is done for the facial expression recognize, in order to estimate the perceptual expression loss.}
\label{fig:arch}
\vspace{-0.5cm}
\end{figure*}

A high-level overview of the architecture is shown in Figure~\ref{fig:arch}. Given a sequence of $K$ RGB frames sampled from an input video $V$, our method reconstructs for each frame $I$ the 3D mesh of the face in FLAME topology, such that the mouth movements and general facial expressions are perceptually preserved. Following the FLAME 3D face model nomenclature, we separate the estimated parameters into two distinct sets:

\paragraph{Rigid \& Identity parameters}
We borrow the coarse encoder from DECA in order to predict independently for each image $I$ in the input sequence the identity $\boldsymbol{\beta}$, neck pose $\boldsymbol{\theta_{neck}}$, albedo $\boldsymbol{\alpha} \in \mathbb{R}^{50}$, lighting $\boldsymbol{l} \in \mathbb{R}^{27}$, and camera $\boldsymbol{c}$. Like EMOCA~\cite{danvevcek2022emoca}, we keep this network fixed through training.

\paragraph{Expression \& Jaw parameters}
The expression $\boldsymbol{\psi}$ and jaw pose $\boldsymbol{\theta}_{jaw}$ parameters that correspond to the input sequence is predicted by an additional ``perceptual" CNN encoder. These parameters explicitly control the mouth expressions and movements under the FLAME framework and therefore should be properly estimated by our approach.
We employ a lightweight MobileNet v2 architecture, but also insert a temporal convolution kernel on its output, in order to model the temporal dynamics of mouth movements and facial expressions in the input sequence.
We selected the aforementioned lightweight option of MobileNet to reduce the computational overhead of our system - contrary to EMOCA- since the existing DECA backbone already uses a resource-demanding ResNet50 model.

In a nutshell, we assume an architecture akin to the one introduced in EMOCA~\cite{danvevcek2022emoca}, with two parallel paths of parameters as described above. 
Nevertheless, our focus is shifted to a very different problem and thus a set of appropriate ``directions" and ``constraints" should be learned through the use of the proposed set of losses, as described in the following section.

\subsubsection{Training Losses}
In order to train the \textit{perceptual} encoder, we use two perceptual loss functions for guiding the reconstruction, along with geometric constraints. 

\paragraph{Perceptual Expression Loss}: The output of the \textit{perceptual} encoder is used along with the predictions of identity, albedo, camera, and lighting in order to differentiably render a sequence of textured 3D meshes, which correspond to the original input video. Then, the input video and the reconstructed 3D mesh are fed into an emotion recognition network (borrowed from EMOCA~\cite{danvevcek2022emoca}) and two sequences of feature vectors are obtained. Then, we apply a perceptual expression loss $L_{em}$, by attempting to minimize the distance between the two sequences of feature vectors. 
Interestingly, even though the emotion recognition network is trained to predict emotions, it can faithfully retain a set of helpful facial characteristics. Therefore, such a loss is responsible for learning general facial expressions, capable to simulate emotions, which promote the realism of the derived reconstruction. Notably, this loss positively affects the eyes, leading to a more faithful estimation of eye closure, frowning actions etc.

\paragraph{Perceptual Lip Movements Loss}
The perceptual expression loss does not retain enough detailed information about the mouth, and as such, an additional mouth-related loss is needed. Instead of relying only on a geometric loss with weak supervision using 2D landmarks, we use an additional perceptual loss, that guides the output jaw and expression coefficients to capture the intricacies of mouth movements. 
\textit{The necessity of such a perceptual mouth-oriented loss is further highlighted by the inaccuracies detected in the extracted 2D landmarks}. For examples of this phenomenon see the Suppl. Material.

For this purpose we use a network that has been trained on the LRS3 (Lip Reading in the Wild 3) dataset~\cite{ma2022visual}. The lip-reading network is the pretrained model provided by Ma et al. \cite{ma2022visual} which takes as input sequences of grayscale images cropped around the mouth and outputs the predicted character sequence. The network has been trained with a combination of Connectionist Temporal Classification (CTC) loss with attention. The model architecture consists of a 3D convolutional kernel, followed by a 2D ResNet-18, a 12-layer conformer, and finally a transformer decoder layer which outputs the predicted sequence (for more details, see~\cite{ma2022visual}).
Our goal here is to minimize the perceptual distance of speech-aware movements between the original and the output image sequences. To that end, we take the differentiably rendered image sequences and subsequently crop the them around the mouth area using the predicted landmarks. Finally, we calculate the corresponding feature vectors $\epsilon_I$ and $\epsilon_R$, from the output of the 2D ResNet-18 of the lip-reading network. We empirically found that features from the CNN output better model the spatial structure of the mouth, while features on the output of the conformer are largely influenced by the sequence context and do not preserve this much-needed spatial structure. 
Examples of this behavior can be found in the Suppl. Material.  
After calculating the feature vectors, we minimize the perceptual lip reading loss between the input image sequence and the output rendered sequence $L_{lr} = \frac{1}{K} \sum^K{d(\epsilon_{I},\epsilon_{R})}$, where $d$ is the cosine distance and $K$ the length of the input sequence.
As a sidenote, initial experiments included an explicit lip reading loss based on the CTC loss over the predicted output of the existing lip reading network, given the original transcription of the sentence. Despite its straightforward intuition, such approach had major downsides apart from the need of the video transcription. First, it had a significant computational overhead since whole sentences should be processed at once. In contrast, the proposed approach simply samples a subset of consecutive frames and tries to minimize the extracted mouth-related features. Moreover, it has proven to be ineffective in practice, suffering from the same behavior as with the features taken from the conformer's output.

\paragraph{Geometric Constraints}
Due to the domain mismatch between the rendered and the original images, although the perceptual losses help retain the high level information on perception, they also tend to create artifacts in some cases. 
This is to be expected; the perceptual losses rely on pre-trained task-specific CNNs that do not guarantee in any way that the input manifold corresponds to realistic images. For example, as we report in Suppl. Material, we can create unrealistic images of distorted facial reconstruction that produce good lip reading results - a typical problem in the adversarial examples topic~\cite{goodfellow2014generative}.
Thus, we guide the training process by enforcing the following geometric constraints:
We regularize the expression and jaw parameters by penalizing their $L^2$ norm with respect to the initial predicted DECA parameters: $||\boldsymbol{\psi} - \boldsymbol{\psi}^{DECA}||^2$ and $||\boldsymbol{\theta}_{jaw} - \boldsymbol{\theta}^{DECA}_{jaw}||^2$.
The aforementioned regularization terms use the estimation from DECA as a ``good" starting point, in the sense that our method should not significantly deviate from DECA parameters, which have been proven to produce artifact-free results in practice. In other words, using such a regularization scheme, we indirectly impose some of the constraints hardcoded by DECA and its training procedure.
We also apply an $L_1$ loss between the landmarks of the nose, face outline and eyes of the 3D model and the predicted landmarks of a face alignment method~\cite{bulat2017far}. For the mouth area we employ a more relaxed $L_2$ relative loss between the intra-distances of mouth landmarks. 
The aforementioned landmark losses comprise an alternative to explicit imposing a geometric loss based on distance between the predicted 2D landmarks of the reconstructed face and the 2D landmarks of the original image. Such a straightforward loss can lead to erroneous reconstruction, as ablation study in supplementary material highlights, since perceptual losses and the 2D landmark loss were often contradicting. Using the proposed version of relative landmark losses achieves retaining the much needed geometric structure of the face without an overly strict constraint that limits the perceptual losses.


Finally, the total loss used for training is then: $L = \lambda_{lr} L_{lr} + \lambda_{em} L_{em} + L{c}$, where $L_c$ includes the previously stated geometric constraints. 

\begin{table*}[!ht]
    \centering
    \begin{tabular}{l|llll|llll|llll}
        ~ & \multicolumn{4}{c|}{LRS3} & \multicolumn{4}{c|}{TCD-TIMIT} & \multicolumn{4}{c}{MEAD} \\ \toprule
        ~& CER & WER & VER & VERW &  CER & WER & VER & VERW  &  CER & WER & VER & VERW   \\ \toprule
RGB & 24.9 & 36.3 & 22.0 & 36.0 & 35.7 & 61.7 & 29.6 & 60.9 & 49.7 & 76.1 & 42.8 & 75.0 \\ \midrule
        DECA & 83.6 & 124.4 & 74.0 & 95.2 & 84.2 & 135.8 & 75.8 & 134.5 & 84.8 & 123.7 & 77.8 & 122.3 \\
        EMOCA & 97.7 & 143.2 & 88.2 & 108.0 & 86.4 & 137.2 & 79.2 & 136.1 & 85.1 & 123.4 & 77.9 & 121.1 \\
        3DDFA\_v2 & 97.5 & 125.0 & 95.3 & 124.5 & 101.8 & 148.7 & 98 & 148.3 & 94.5 & 124.8 & 90.2 & 124.0 \\ 
        DAD & 84.1 & 111.2 & 78.2 & 110.4 & 87.3 & 135.1 & 81 & 134 & 86.0 & 119.7 & 79.9 & 118.2 \\ \midrule
        ours & \textbf{67.6} & \textbf{93.3} & \textbf{60.8} & \textbf{92.2} & \textbf{75.6} & \textbf{120.2} & \textbf{67.1} & \textbf{119.5} & \textbf{79.6} & \textbf{114.9} & \textbf{72.4} & \textbf{113.6} \\ \bottomrule
    \end{tabular}
    \caption{Lipreading results on the LRS3-test, TCD-TIMIT and MEAD datasets (network trained on LRS3-train set). For all metrics, lower is better (error rates). Our method significantly outperforms all other 3D reconstruction methods. ``RGB" denotes results on the original video footage. }
    \label{tab:obj}
\end{table*}


\subsection{Training Details}
We train our network on the Lip Reading Sentences 3 (LRS3) dataset~\cite{afouras2018lrs3}. This is the largest publicly available dataset for lip reading in the wild. We use the official \textit{trainval} (31,982 utterances) set for training and validating our model. We train using Adam optimizer with starting learning rate 5e-5, reducing the rate 5-fold at $50,000$ iterations. We use a sequence length of $K=20$ and batch size 1. Source code is provided in the supplementary material.

\section{Experiments}
We evaluate our method both qualitatively and quantitatively, following a similar evaluation procedure with ~\cite{danvevcek2022emoca}. For evaluation we use the following datasets:

\textbf{LRS3}~\cite{afouras2018lrs3}: The test set of LRS3 (1,321 utterances).

\textbf{MEAD}: This is a recent dataset~\cite{kaisiyuan2020mead} containing 48 actors (28M, 20F) from multiple races uttering sentences from TIMIT~\cite{garofolo1993darpa} in 7 basic emotions (happy, angry, surprised, fear, sad, disgusted, contempt) plus neutral and 3 different levels of intensity. The whole dataset includes 31,059 sentences. We randomly sampled 2,000 in order to create a test set, stratifying for subject, emotion, and intensity level. 
\textbf{TCD-TIMIT}~\cite{harte2015tcd}: This corpus includes 62 english actors reading 6913 sentences from the TIMIT~\cite{garofolo1993darpa} corpus. We use the official test split for evaluation.

We compare our method with the following recent state-of-the-art methods on 3D facial reconstruction:
\textbf{DECA}~\cite{feng2021learning}, \textbf{EMOCA}\cite{danvevcek2022emoca}, \textbf{3DDFAv2}~\cite{3ddfa_cleardusk}, and \textbf{DAD-3DHeads} which uses direct 3D supervision from the large-scale annotated DAD-3DHeads~\cite{dad3dheads} dataset. Note the lack of recent methods for 3D reconstruction in video. As a result, in order to reconstruct the input video, we apply each method per frame. Especially for 3DDFAv2, we apply temporal smoothing as provided by the official implementation. For all methods we use the official implementation. \textbf{Additional results and visualizations are provided in the video of the Supp. Material.}

\subsection{Objective Results}
The difference between a reconstructed 3D facial expressions and the corresponding ground truth can be dominated by errors corresponding to the identity of the person, evaluating using a geometric criterion does not necessarily not correlate well with human perception of expression and mouth movements~\cite{danvevcek2022emoca}. As a result, we evaluate the methods objectively in terms of lip reading metrics by apply a pretrained lipreading network on the output rendered images. To remove bias, we use a \textit{different architecture and pretrained lipread model} for evaluation than the one used for the lipread loss, which is based on the Hubert transformer architecture, called AV-HuBERT~\cite{shi2022avhubert,shi2022avsr}. We report the following metrics: Character Error Rate (CER) and Word Error Rate (WER), as well as Viseme Error Rate (VER) and Viseme-Word Error Rate (VWER), obtained by converting the predicted and ground truth transcriptions to visemes using the Amazon Polly phoneme-to-viseme mapping~\cite{borntrager_2015}. Results are presented in Table~\ref{tab:obj}. Our method achieves much lower CER, WER, and VER scores compared to the other methods, both in the LRS3 test set, as well as in the cross-dataset evaluations on TCDTIMIT and MEAD. In the same Table we also include results on the original video footage, which showcase the domain gap ``problem" (more information about this in Discussion section) of the used lip reading systems: the pre-trained models have been trained to the initial images without the possible visual degradation introduced by the rendering procedure.
Nonetheless, our method provides notable boost in lip reading performance, despite missing key features such as tongue and teeth, by properly encoding speech-aware features.



\begin{figure*}[t]
\centering
\includegraphics[width=\textwidth]{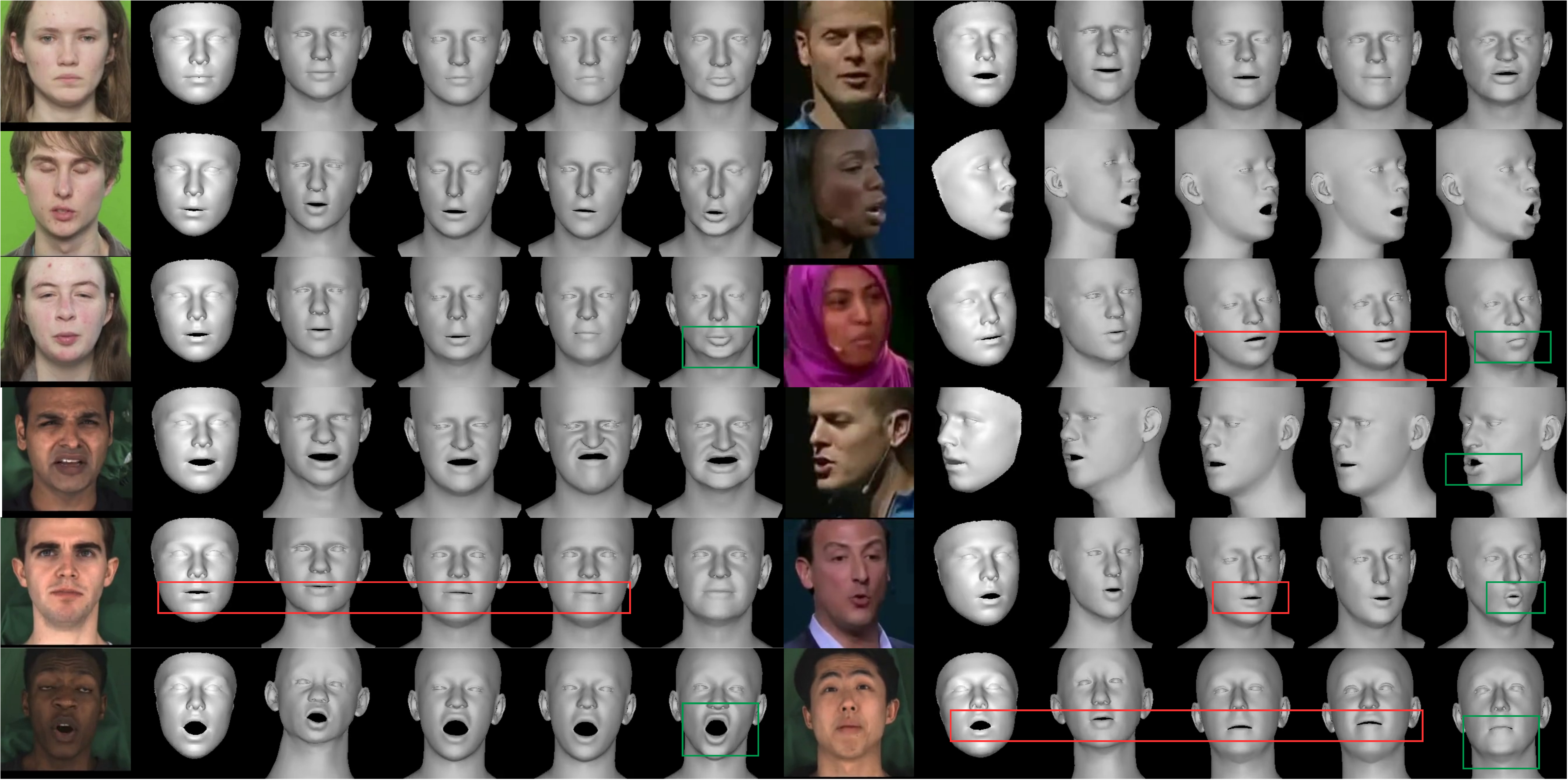}

\caption{Visual comparison with other methods on the MEAD, TCDTIMIT, and LRS3 datasets. Note that our method is only trained on the LRS3 train test. From left to right: original footage, 3DDFAv2~\cite{3ddfa_cleardusk}, DAD~\cite{dad3dheads}, DECA~\cite{feng2021learning}, EMOCA~\cite{danvevcek2022emoca}, ours. We also highlight with red boxes some erroneous results, and with green boxes some examples of retaining the original mouth formation.} 
\label{fig:broken}
\vspace{-0.5cm}
\end{figure*}


\subsection{Subjective Results}
To assess the realism and perception of the 3D reconstructed faces in humans we have designed and conducted two web user studies~\cite{kritsis2022danceconv}. In order to mitigate any intra-dataset bias that might arise from training on the LRS3 trainset and showing users video from its test set, for these studies, we used only videos from the MEAD and TCD-TIMIT dataset.

\paragraph{First Study: Realism of Articulation}
For this study, we selected a preference test design, by showing users pairs of 3D reconstructed faces, alongside the original footage, and asking them to select the most realistic one in terms of mouth movements and articulation. We created a question bank consisting of 30 videos from the MEAD dataset (21 emotional videos for each level of intensity and emotion and 9 neutral), and 10 videos from the TCD-TIMIT dataset and performed 3D reconstruction using the previously stated 5 methods (DAD, DECA, EMOCA, 3DDFAv2 and ours). Then, users were presented with two randomly ordered reconstructed faces, each alongside the original footage, and were asked to choose the most realistic one in terms of mouth movements and articulation. Each user answered 28 randomly sampled questions from the bank (7 questions for each pair - ours vs the others), and a total of 34 users completed this study.

The results of this study can be seen in Table~\ref{tab:study1}. We can see that our method is significantly preferred to all other methods ($p<0.01$ with binomial test, adjusting for multiple comparisons using the Bonferroni method). 3DDFAv2\cite{3ddfa_cleardusk} was method least preferred with DECA and EMOCA following.  The results clearly highlight the importance of the proposed method from the speech-aware perspective and how humans favorably perceive the reconstructed mouth movements.

\paragraph{Second Study: Lip Reading}
In the second study, users (disjoint set of participants compared to the first study) were presented with a muted video of a person speaking a specific single word in the form of a 3D talking head reconstructed from one of the compared methods and then were asked to find which word is being said from 4 different alternatives (multiple choice). For this, we cropped 40 single words from the MEAD and TCD-TIMIT datasets, covering different visemes, and presented each user with a random subset of 30 words (6 words for each method in each questionnaire). A total of 31 users completed this study. Classification results are shown in Table~\ref{tab:study2}. 
It is interesting to see that our method achieves similar scores to EMOCA and DAD, even though EMOCA did not explicitly model the mouth jaw. This points to the fact that even though our method is significantly more realistic in terms of articulation, as the first user study supports, there are cases in which humans are not able to correctly identify the word or even cases when semi-erroneous articulation, e.g. unrealistically exaggerated as in the case of EMOCA, can be sufficient for distinguish specific words. A per word analysis with visual examples is provided in the supplementary material.Despite the low accuracy, our system seems to marginally outperforms the compared SoTA methods in the challenging task of lip-reading performed by non-experts.



\begin{table}[]
\begin{tabular}{l|l|l|l|l}
     & DECA   & EMOCA  & 3DDFAv2 & DAD    \\\toprule
Ours & \textbf{201}/37 & \textbf{185}/53 & \textbf{218}/20  & \textbf{150}/88 \\ \bottomrule
\end{tabular}
\caption{Preference results of the first subjective study. Our method is \textbf{significantly} ($p<0.01$ with binomial test after adjusting for multiple comparisons) more realistic in terms of mouth movements and articulation.}
\label{tab:study1}
\end{table}

\begin{table}[]
\begin{tabular}{l|l|l|l|l}
   Ours  & DECA   & EMOCA  & 3DDFAv2 & DAD    \\\toprule
   \textbf{47.3}\% & 38.7\% & 45.2\% & 23.6\%  & 46.7\% \\ \bottomrule
\end{tabular}
\caption{Classification accuracy obtained from the second user study (word-level lip-reading). More detailed results are included in the Suppl. Material.}
\label{tab:study2}
\end{table}

\subsection{Ablation study}

\begin{figure}[t]
\centering
\includegraphics[width=.5\textwidth]{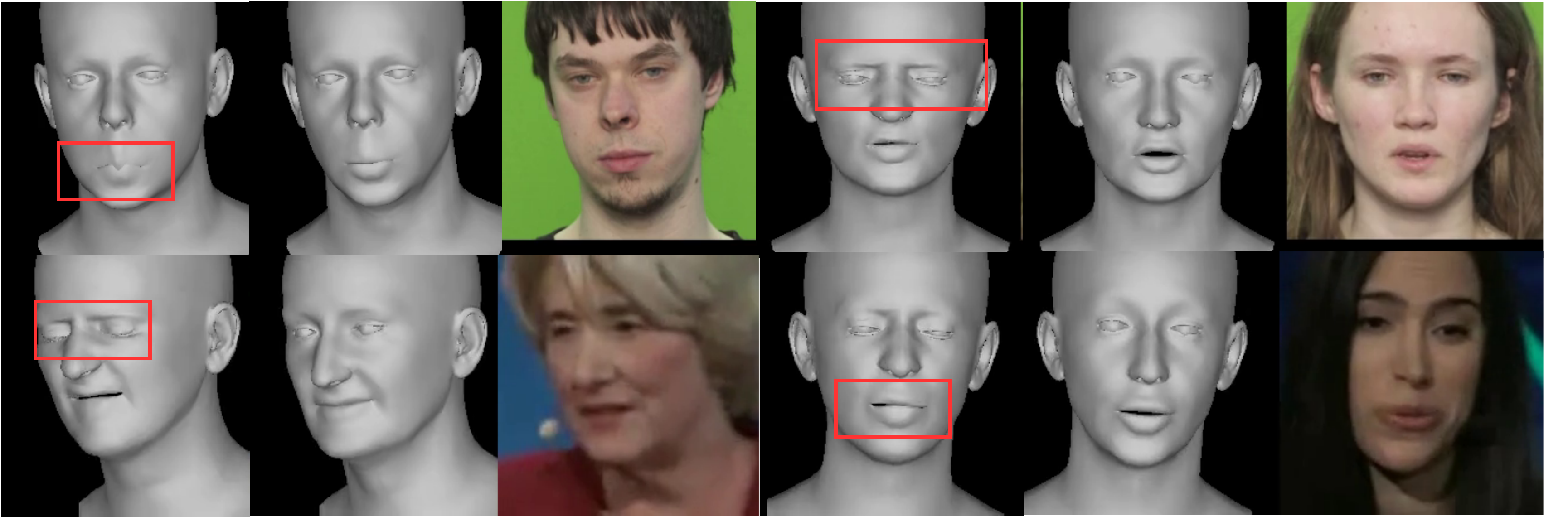}

\caption{Training of the perceptual encoder without (left) and with (middle) geometric constraints based on 2D landmarks. Omitting geometric constraints from the rest of the face leads to appearance of artifacts in the eyes and nose in some cases, while completely omitting 2D information from mouth landmarks can lead to some failure cases in the mouth area. Please zoom in for details.}
\label{fig:ablation1}
\vspace{-0.5cm}
\end{figure}

In Fig.~\ref{fig:ablation1} we show results of training the network with and without the geometric constraints from landmarks. We can see that in some cases, completely removing geometric constraints and training only with perceptual losses leads to artifacts around the eyes, nose and mouth shape.

Finally, in Fig.~\ref{fig:broken} we also present multiple visual comparisons from the 3 datasets with the four other methods.

\section{Discussion}
Our method has introduced a significant step towards creating truly realistic 3D talking heads, as it has been shown by our extensive objective and subjective evaluation against other SoTA methods. It is important to note that our method even outpeforms DAD, which was trained with 3D annotated data on a large-scale dataset. It should also be pointed out, as it is also evident in Figure~\ref{fig:broken} that the lipread loss, not only retains the motions and shape of the mouth, but it also makes it more distinct in the rendered mesh. It becomes apparent that in order to achieve realism in terms of speech, we need to opt for more perceptual losses. This has also been done in previous methods regarding the emotional expression~\cite{danvevcek2022emoca} as well as 3D shape~\cite{feng2021learning,zielonka2022mica}. 
Note also that training with our lipread loss does not require any kind of text transcriptions or the corresponding audio.
Moreover, even though our method is trained in speech videos according to the proposed lipread loss, it can be used to model arbitrary mouth movements, non-related to speech. This generalization property stems from the fact that we train to perceptually simulate mouth movements and thus the encoded mouth features are not necessarily correspond to speech-related movements.

\paragraph{Limitations}
We point out that the results of the objective evaluation on CER and WER, remain much higher compared to the original footage. This is of course, among others a problem of the different domain of the rendered images compared to the ground truth. The absence of teeth and tongue is also important, since they play a large role in the detection of specific types of phonemes/visemes such as alveola and dental consonants. This domain adaptation problem has not been addressed in this work, since our approach works well in practice, but it remains a hindrance to unleashing the full potential of the described losses.
This domain problem also includes the perceptual losses. Both perceptual losses make the assumption that the original images and the rendered ones belong to the same visual ``domain". Nonetheless, there is indeed a realism/domain gap between these two feature spaces that may lead to inconsistencies; this is why we needed to have relative landmarks. As a result, the loss of landmarks and the lipread loss sometimes compete against each other: on one hand, lip reading tries to improve the perception of the talking head while landmarks, if not detected accurately tend to reduce the realism. On the other hand, we have observed that from a threshold and lower, reduction of lip read loss tends to create artifacts; which is why we need the constrains from landmarks to retain the realism of the facial shape. 
In addition, although our method includes a loss borrowed from EMOCA~\cite{danvevcek2022emoca}, in order to retain the facial expressions outside the mouth (e.g. in eyes), since it was trained only on the LRS3 dataset (which does not include emotional samples) the results in some cases tend to not include the intensity of emotion present in EMOCA. Furthermore, note that while DECA and EMOCA included detailed refinement, by calculating a detailed UV displacement map, which modeled person specific details such as wrinkles, our method does not include this step.Finally, while as we have already stated our method does not need text transcriptions or audio, we believe that these modalities, if present in the dataset, could be leveraged in order to improve the total perception.


\paragraph{Societal Impact}
While we do not believe that the method described in this text can have any direct negative applications, we are aware that the end goal of this method, which is fully realistic 3D reconstruction of human talking heads can be used negatively, as demonstrated recently in deepfake technology. As a result, we believe that researchers active on the field of 3D face reconstruction and synthesis should also at the same time explore methods that accurately detect fake 3D reconstructions~\cite{rossler2018faceforensics}.

\section{Conclusion}
We have presented the first method for visual speech-aware perceptual reconstruction of 3D talking heads. Our methods does not rely on text transcriptions or audio; on the contrary we employ a ``lipread" loss, which guides the training process in order to increase the perception of mouth. Our extensive subjective and objective evaluations have verified that the results of 3D reconstruction are significantly preferred to counterpart methods which rely only on geometric losses for the mouth movements, as well as to methods that use direct 3D supervision. We believe that we have performed an important step towards reconstructing truly realistic talking heads, by focusing not on the purely geometric-based aspect of things, but also on the perception from humans.

\clearpage
{\small
\bibliographystyle{ieee_fullname}
\bibliography{egbib}
}

\clearpage

\section*{Supplementary Material}

\maketitle
\appendix

\section{Inaccuracies of 2D landmarks}
\begin{figure}[h]
\centering
\includegraphics[width=.45\textwidth]{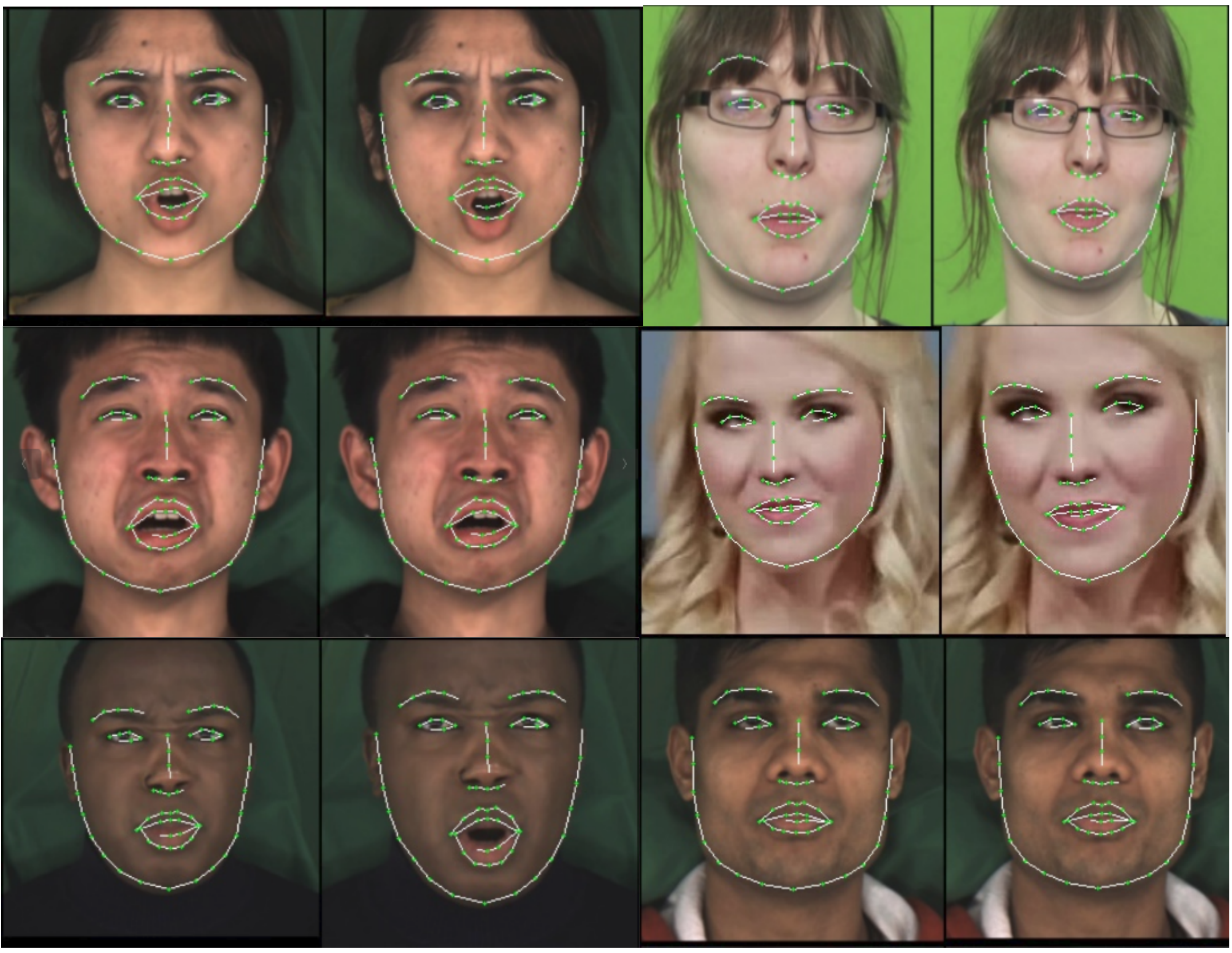}

\caption{Examples of inaccuracies in 2D landmark detection. Notice how especially on the right column the face alignment has not accurately predict mouth closure which is of vital important for realistic perception of bilabial consonants (/p/, /m/, /b/).}
\label{fig:lds}
\end{figure}

As we stated in the main text, the necessity of a perceptual visual-speech aware mouth loss stems from the inaccuracies that we have observed in face alignment methods. In traditional 3D reconstruction in-the-wild, 2D landmarks from a face alignment method are used for weak supervision~\cite{danvevcek2022emoca,3ddfa_cleardusk,feng2021learning} to make up for the lack of ground truth 3D. However, 2D landmarks, especially around the mouth prove to be a poor guide for reconstructing accurate and perceptually realistic mouth movements. We show examples of inaccurate 2D landmarks in Figure~\ref{fig:lds}.

Nonetheless, apart from inaccurate prediction of 2D landmarks in several cases, weak 2D supervision for dense modeling is ill-posed, especially for the lip area which can assume extremely diverse formations. A final remark on this subject also should be the fact that while some lip landmarks (in the most commonly used template of 68 facial landmarks) such as the lip corners have a semantic meaning, intermediate lip landmarks have an intrinsic ambiguity in their definition and present significant variance across different annotators~\cite{sagonas2016300}.

\section{Ablation study on lipreading features and CTC loss}
\subsection{ResNet18 vs Conformer features}
As mentioned in Section 3.2.1 of the main text, we selected features from the ResNet18 output of the lipreading network instead of latter features from the output of the conformer. We present here an ablation study between these two features. For this ablation study, in order to study the immediate effect of different features, we directly optimize the initial estimation of DECA~\cite{feng2021learning} expression $\boldsymbol{\psi}$ and jaw pose $\boldsymbol{\theta}_{jaw}$ parameters using the lipread and regularization losses: $L = \lambda_{lr} L_{lr} + \lambda_{\psi} L_{\psi} + \lambda_{\theta_{jaw}} L_{\theta_{jaw}} $ where $\lambda_{lr}=4$, $\lambda_{\psi}=1e-3$, and $\lambda_{\theta_{jaw}}=200$. We avoided using the relaxed geometric loss from landmarks in this study in order to see the full effect of the different features.  

The results of this ablation are in Figure~\ref{fig:ablation1}. For each image sequence sample in the Figure we show in the top row the original footage, in the 2nd row the initial estimate of DECA~\cite{feng2021learning}, in the 3rd row the result of optimizing the lipread loss using Conformer features, and the last row optimizing the lipread loss using ResNet18 features. Although the conformer preserves useful information for the mouth area, there is not a strict visual correspondence with the original images, because the features are largely affected by the sequence context. On the other hand, features from ResNet18 retain the spatial structure and strict correspondence and are more suited to use for the perceptual lipread loss.

\subsection{CTC loss and adversarial examples}
\begin{figure}[t]
\centering
\includegraphics[width=.48\textwidth]{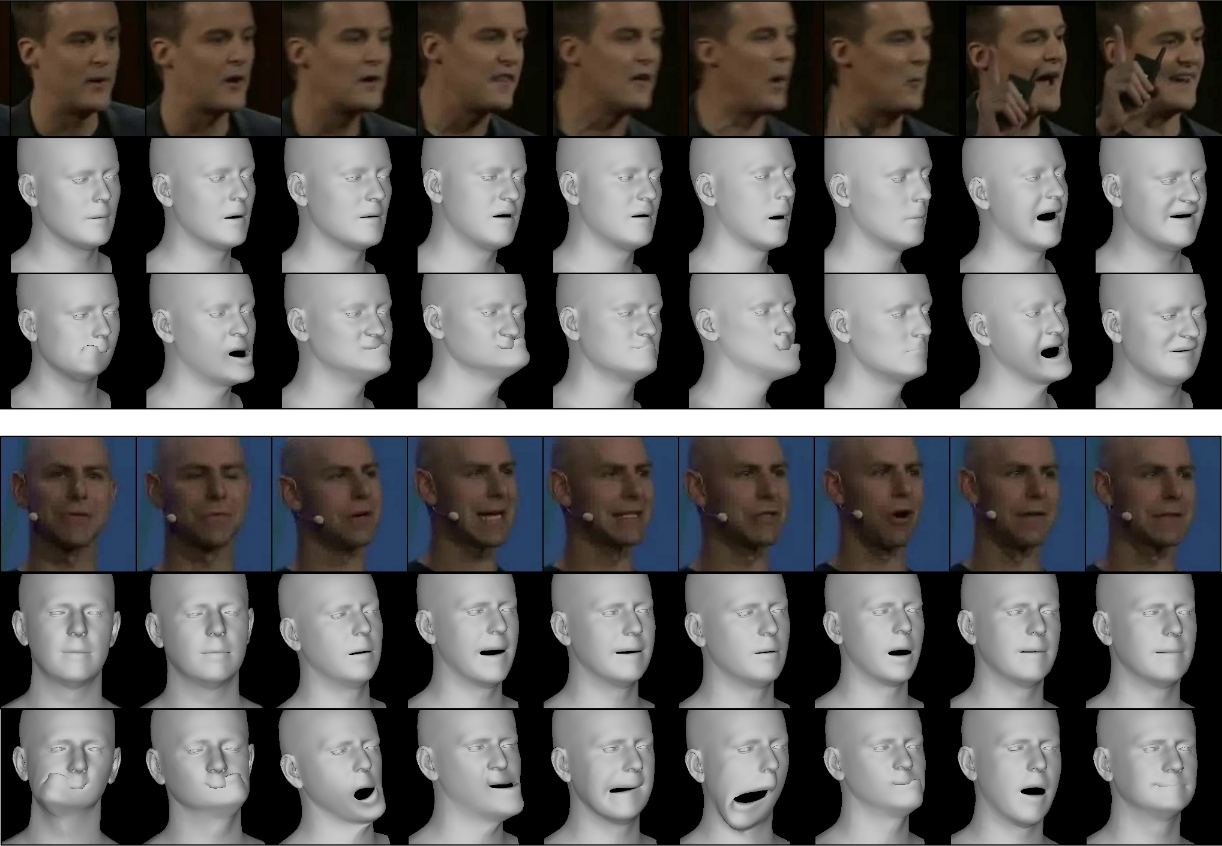}

\caption{Two examples of adversarial attacks using the CTC loss. Middle row shows sampled frames from the original predicted sequence by DECA~\cite{feng2021learning} of two sentences with starting CER (character error rate) around 0.90, while the third row shows completely distorted examples which however achieve near-perfect CER.}
\label{fig:adv}
\end{figure}

We also considered in our initial experiments leveraging text transcriptions and enforced a Connectionist Temporal Classification (CTC)~\cite{graves2006connectionist} loss on the text prediction of the lipreader. Apart from some straightforwards downsides such as the computational overhead of processing whole sentences at once this loss, and requirement of text transcriptions, not only did this loss not retain any spatial structure, but also led to emergence of completely distorted facial reconstructions that achieved a perfect lip reading recognition - a common phenomenon found in adversarial attacks~\cite{goodfellow2014generative,akhtar2018threat}. We showcase this behavior in Figure~\ref{fig:adv}.

\begin{figure}[htb]
\centering
\includegraphics[width=.44\textwidth]{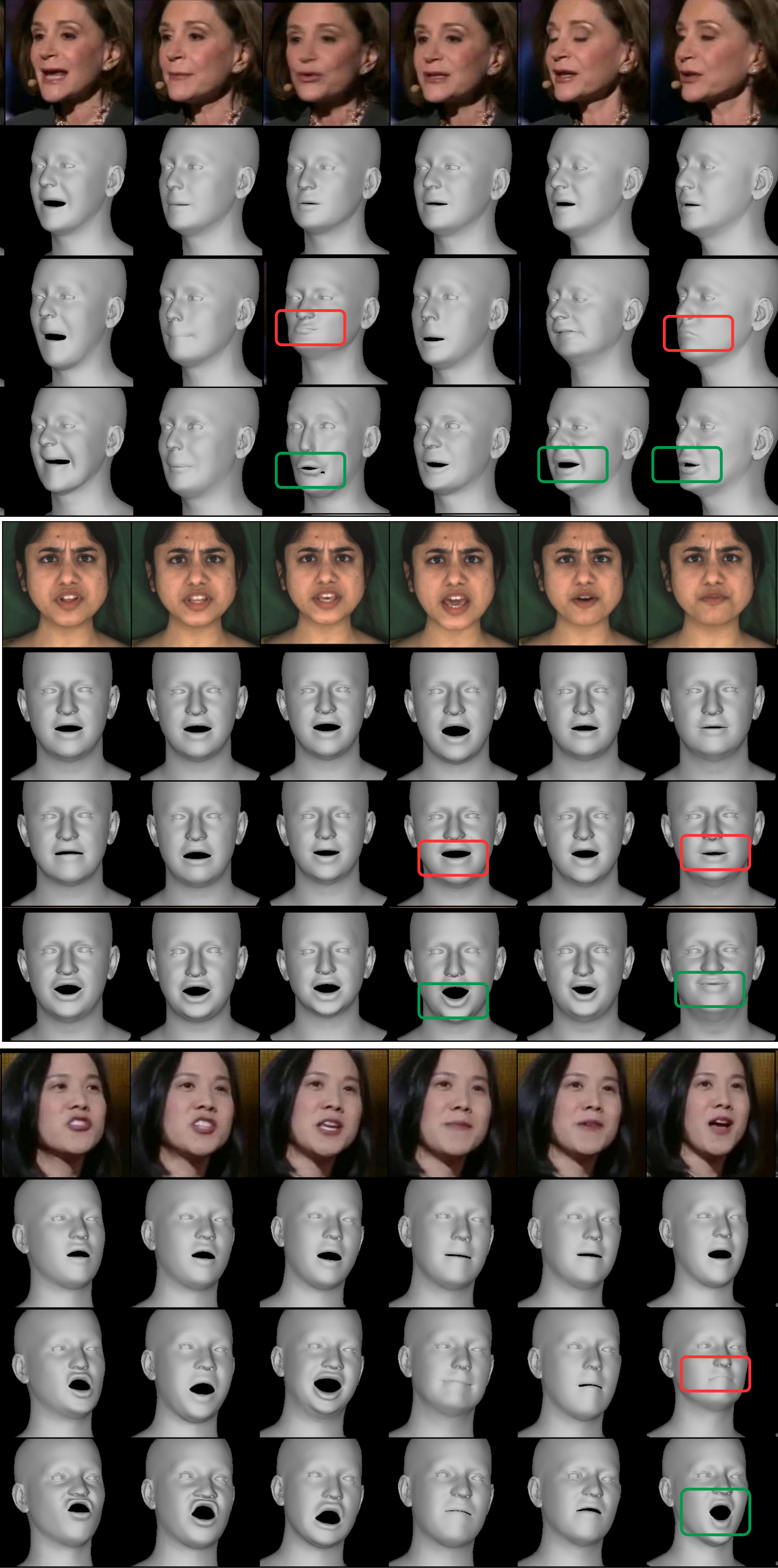}

\caption{Comparison between features from the ResNet18 network of the lipread versus features from the conformer level. For each example, the top row shows the original footage, 2nd row the prediction of DECA~\cite{feng2021learning}, the third row the result of optimizing the lipread loss using conformer features, and the last row using the ResNet features. While features from the conformer do improve the mouth area, they do not have a strict visual correspondence with the corresponding original images. On the other hand, features from ResNet better retain spatial information about the mouth structure. }
\label{fig:ablation1}
\end{figure}

\section{Details on geometric constraints and loss function for the perceptual encoder}
In Section 3.2.1 apart from the perceptual expression and lip movements loss we also briefly mentioned our geometric constraints loss $L_c$ which is used to guide the optimization process, to mitigate problems that arise from the domain gap between the input and rendered images. This geometric constraints loss includes the $L^2$ norm of the expression $\psi$ and jaw pose $\theta_{jaw}$ parameters with respect to the initial estimate of deca: $L_{\psi} = ||\boldsymbol{\psi} - \boldsymbol{\psi}^{DECA}||^2$ and $L_{\theta_{jaw}}= ||\boldsymbol{\theta}_{jaw} - \boldsymbol{\theta}^{DECA}_{jaw}||^2$. 

In addition, we also apply an $L_1$ loss between the predicted and original landmarks (obtained with face alignment~\cite{bulat2017far} of the \textbf{nose}, \textbf{eyes} and \textbf{face outline}: $L_{n} = ||\boldsymbol{E}_{r}-\boldsymbol{E}_{gt}||$, where $\boldsymbol{E}_r$ are the predicted and $\boldsymbol{E}_{gt}$ the original landmarks. For the mouth however, we employ a more relaxed constraint by using the intra-distances of mouth landmarks instead of the direct values: $L_{m} = ||\boldsymbol{D^{m}_{r}} - \boldsymbol{D^{m}_{gt}}||^2 $, where $\boldsymbol{D^{m}_{r}}$ are the distances between pairs of the predicted mouth landmarks while $\boldsymbol{D^{m}_{gt}}$ are the distances of pairs of original mouth landmarks. 

\begin{figure}[h]
\centering
\includegraphics[width=.48\textwidth]{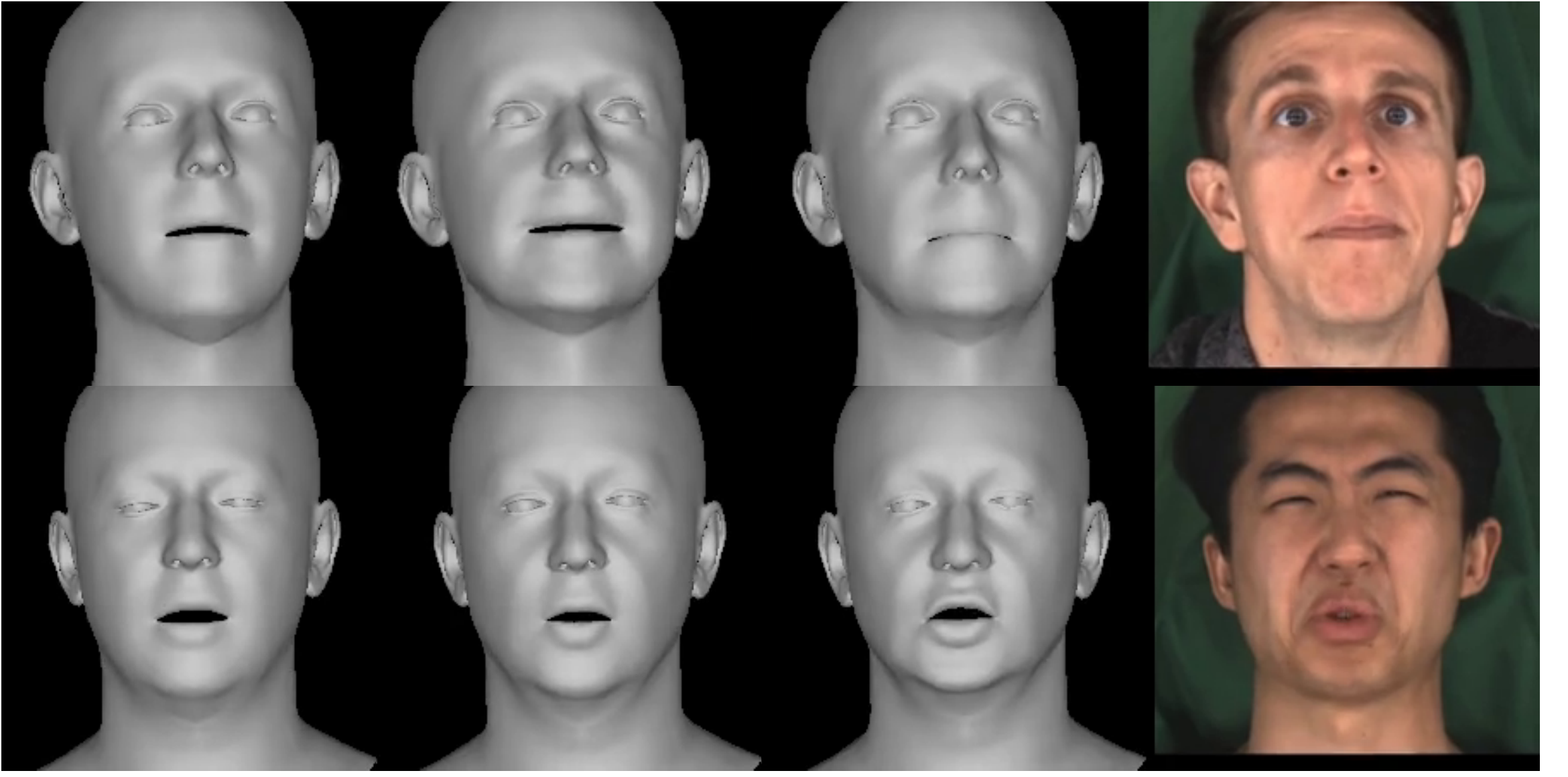}

\caption{Ablation between using absolute position of mouth landmarks or relative intra-mouth distances. The first column is the initial estimate of DECA, the second column the predicted reconstruction of a model trained with an $L_1$ loss imposed on the mouth landmarks as well, and the third column a model trained with a more relaxed loss on the mouth using the intra-mouth distances. Strict mouth landmark losses erroneously guide the output to resemble DECA, while the relaxed constraints leave enough freedom to the perceptual loss to accurately capture the formation of lips.}
\label{fig:rel}
\end{figure}

We use this more relaxed version because a straightforward loss between the predicted and original landmarks is more strict and can lead to erroneous reconstructions, since perceptual losses and the 2D landmark loss can be contradicting. For example, observe Fig.~\ref{fig:rel}. In this example, the left column shows the initial 
estimate of DECA, the middle column the predicted reconstruction of a model trained with an $L_1$ loss imposed on the mouth landmarks as well, and the third column a model trained with a more relaxed loss on the mouth using the intra-mouth distances. As it can be seen, strict landmark losses guide the result to resemble DECA. On the other hand, relative losses are less strict, and the model accurately predicts the mouth structure. 
Concluding, the perceptual encoder is trained with the following criterion:

\begin{equation*}
 L_{pc} = \lambda_{lr} L_{lr} + \lambda_{em} L_{em} + \lambda_{\psi} L_{\psi} + \lambda_{\theta_{jaw}} L_{\theta_{jaw}} + \lambda_{{n}} L_{{n}} + \lambda_{{m}} L_{m}     
\end{equation*}

where $\lambda_{lr}=2$, $\lambda_{em}=0.5$, $\lambda_{\theta_{jaw}}=200$, $\lambda_{{n}}=50$, $\lambda_{{m}}=50$. 
Note that especially for the weight $\lambda_{\psi}$, we selected a nonlinear weighting scheme:

\begin{equation}
\lambda_{\psi} = 
\left\{
    \begin{array}{lr}
        1e-3, & \text{if } L_{\psi} < 40\\
        2e-3, & \text{if }  L_{\psi} > 40
    \end{array}
\right\}
\end{equation}
which we found in practice to work better than a traditional fixed weight. 
The motivation behind this nonlinear tweak of the regularization term is to impose stricter constraints after an empirical threshold, since we have observed that the necessity to continuously minimize the reported perceptual losses may lead to artifacts. Even though this modification does not significantly affects the procedure, we found it effectively reduced specific artifacts.

\section{Failure Cases}
\begin{figure}[h]
\centering
\includegraphics[width=.48\textwidth]{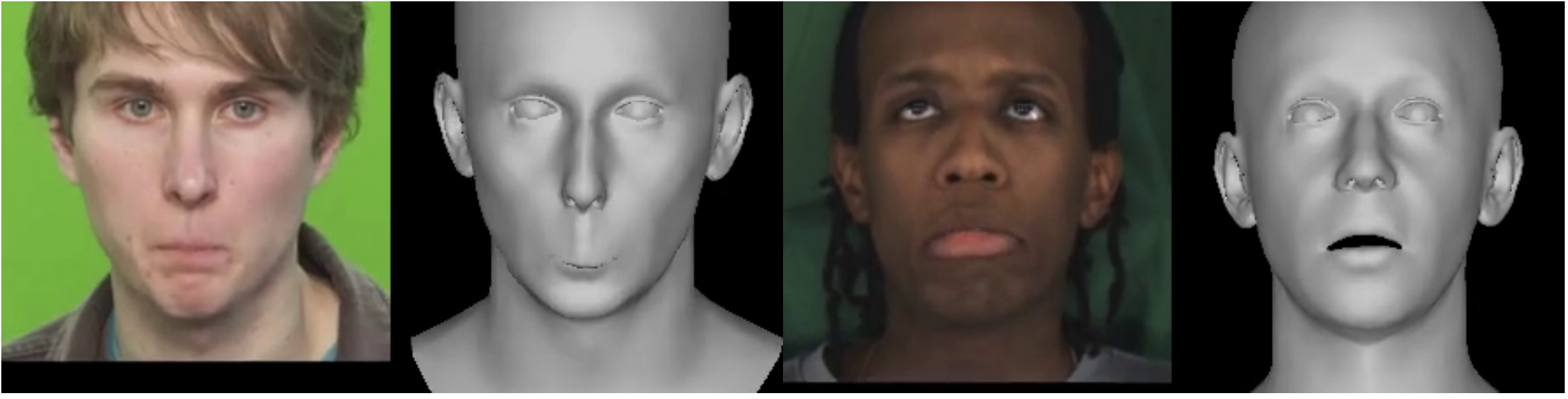}

\caption{Examples from failure cases of our model. The domain gap problem can still cause some mouth artifacts, even when guided by our geometric constraints. Note also that any failed results of the lipread network propagate to our 3D reconstruction method as well.}
\label{fig:fails}
\end{figure}

Finally, we also include in Figure~\ref{fig:fails} two examples of erroneous mouth reconstructions of our model. In the first example there is an artifact in the mouth area, while in the second example, the reconstructed 3D shape has erroneously an open mouth. We believe that there are two major factors which can negatively affect our method. First, while our geometric relative constraints have greatly alleviated the domain gap problem in the perceptual losses, we can still find samples where this problem has created some minor artifacts. Second, since the perceptual loss itself originates from a neural network, failure cases of the lipread loss propagate to our 3D reconstruction model. 


\begin{figure*}[h]
\centering
\begin{tabular}{cc}
 \includegraphics[width=.45\textwidth]{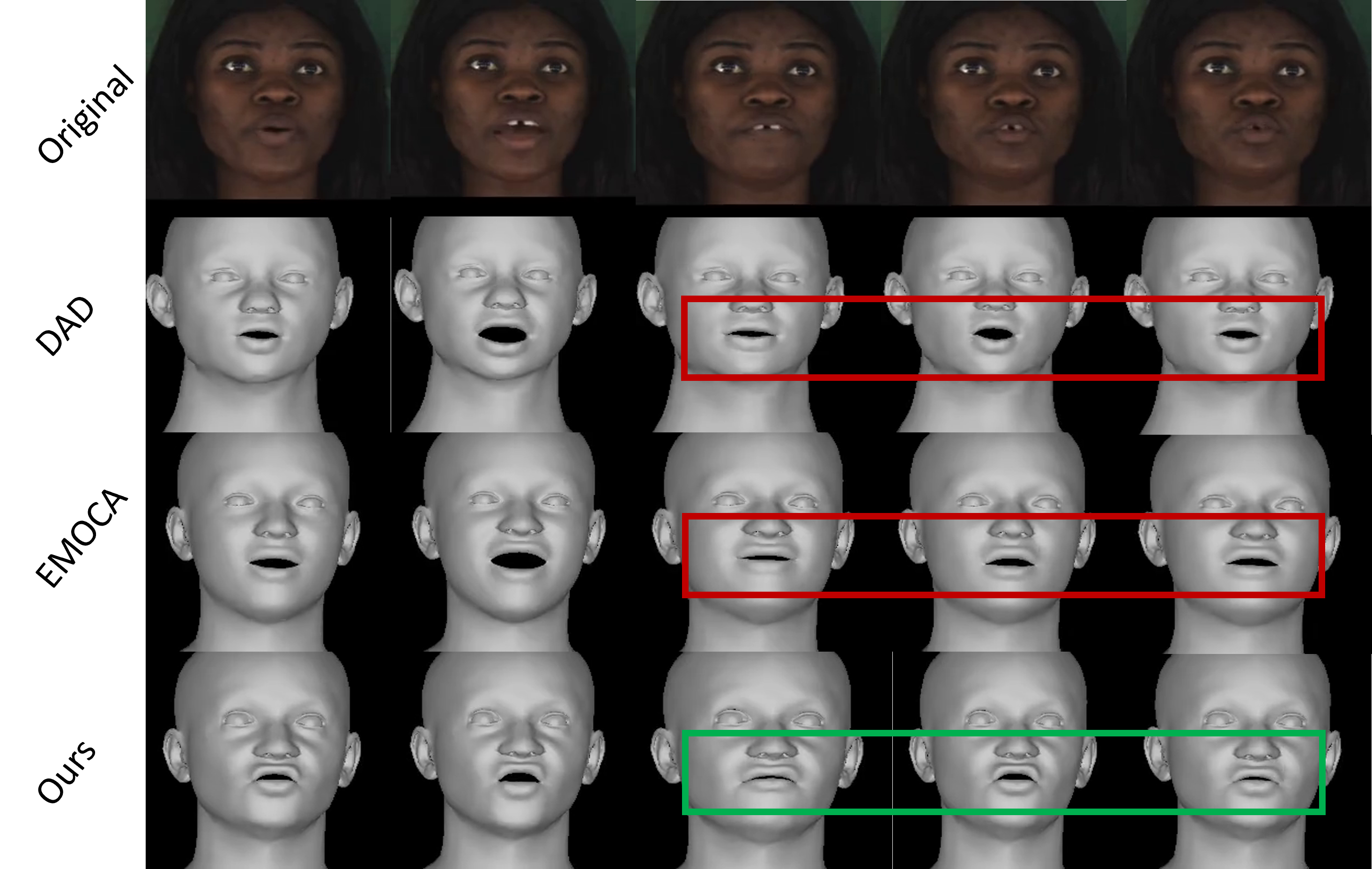} &
 \includegraphics[width=.45\textwidth]{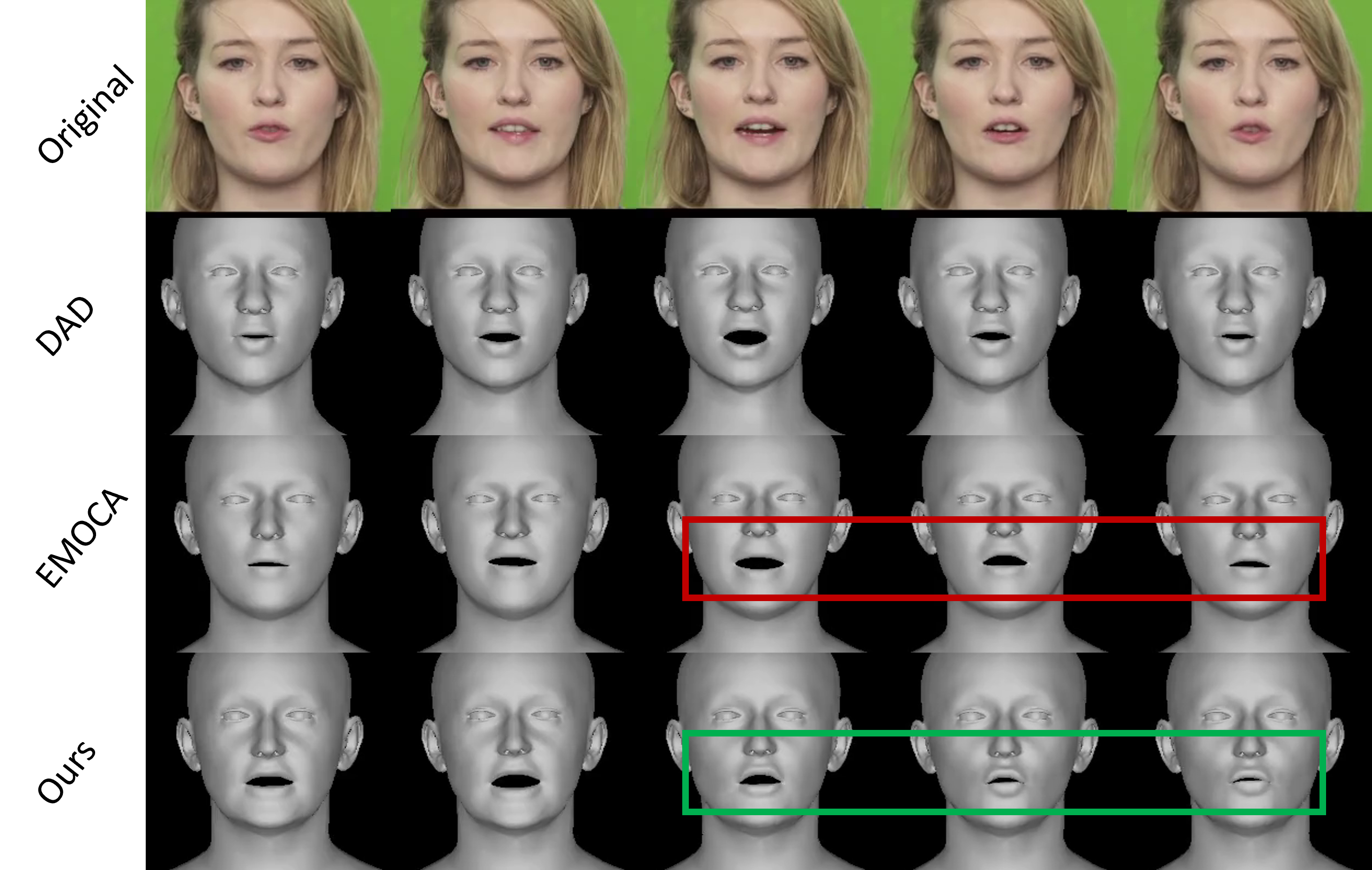} \\
 (a) PERFUME & (b) NARROW \\
  \includegraphics[width=.45\textwidth]{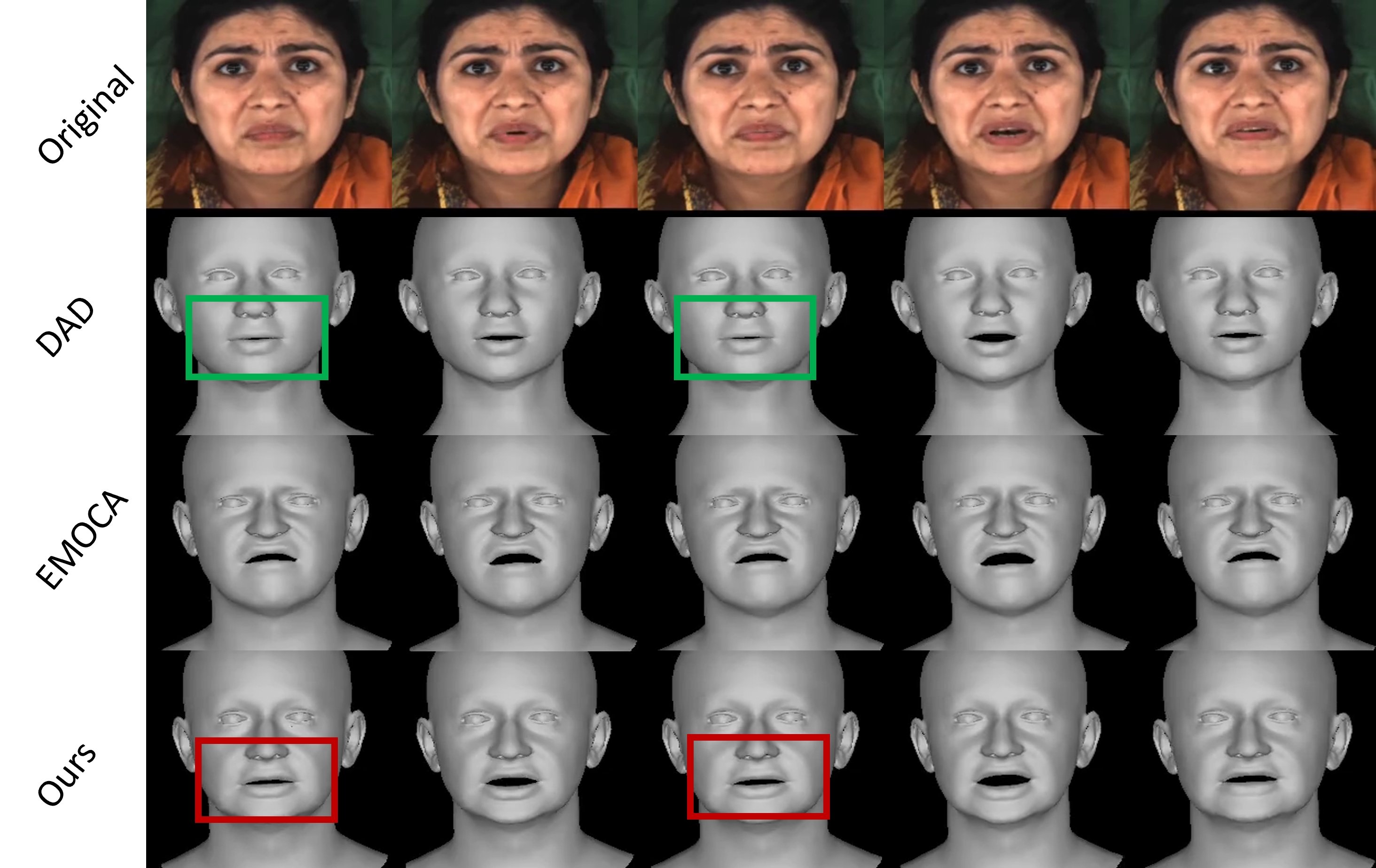} \\
 (c) PEOPLE \\
\end{tabular}
\caption{Three example words from the second user study (lip reading). We show our method against DAD and EMOCA. In PERFUME and NARROW, our method accurately predicts the rounded mouth formations. In the third case of PEOPLE, we see a failure case, where the bilabial consonant /p/ which corresponds to closed mouth was not predicted accurately. Note how also in PERFUME, our method accurately depicts /f/ in the third frame.}
\label{fig:word-examples}
\end{figure*}

\begin{table*}[]
\resizebox{\linewidth}{!}{
\begin{tabular}{l|ccccc|ccccc}
\footnotesize
& PLACE & PEOPLE & WITHDRAW	& AROUND & CONSIDERABLE & PROBLEM &	WHATEVER & NARROW & AUTHORIZED & PERFUME
\\\toprule
DAD & 60 & \textbf{100} & 71 & \textbf{100} & 67 & 25 & 50 & 78 & 50 & 50 
\\
EMOCA  & \textbf{67} & 20 & \textbf{100} & 33 & \textbf{100} &  17 & 40 & 67 & 0 & 67 
\\
DECA   & 20 & 50 & 17 & 62 & 50 & 0 & 50 & 0 & 25 & 50
\\
3DDFAv2 & 33 & 0 & 20 & 50 & 0 & 20 & 40 & 0 & 38 & 40
\\
Ours & 25 & 57 & 40 & 40 & 67 & \textbf{67} & \textbf{71} & \textbf{89} & \textbf{100} & \textbf{100}
\\ \bottomrule
\end{tabular}
}
\caption{Per-word recognition results for the second user study, including all considered SoTA methods. We report indicative cases of failure (first five columns) and success (last five columns) of our approach compared to the other methods.}
\label{tab:study2-analysis}
\end{table*}

\section{Analysis of Second User Study}
As we showed in the second user study of Section 4.2 (Lip Reading study), even though our method is significantly more realistic in terms of mouth motion (as pointed out by the first user study) compared to other methods, it achieved a marginally better performance with EMOCA~\cite{danvevcek2022emoca} and DAD~\cite{dad3dheads}. Here, we show a per-word in depth evaluation of the results of the second user study. 
Specifically, we report the recognition accuracy in Table~\ref{tab:study2-analysis} for five indicative cases where our method under-performs and also five cases where our method outperforms competition. In addition, in Figure~\ref{fig:word-examples} we also show example video reconstructions of three words: PERFUME, NARROW, and PEOPLE. In the first two words our method had a significantly higher recognition accuracy, while in the last one, DAD performed better. As we can see from the visual comparison in ~\ref{fig:word-examples}(c), our method in this specific case failed to accurately capture the closed mouth formations that correspond to the bilabial consonants /p/. On the other hand,in the the first two words our method accurately captures the mouth formations for the rounded vowels /o/ and /u/ in contrast with EMOCA and DAD. Note how also in PERFUME, our method accurately depicts /f/ in the third frame.

In the study there were also cases where the majority of the methods perform well due to the very distinct pronunciation of the words (e.g. ``BALEFUL" and ``UMBRELLA") and cases where all methods considerably under-perform (e.g., ``GREASY", ``SURRENDER") due to subpar reconstruction and ``difficult" alternative words (e.g., ``SURRENDER" was mostly confused with the alternative choice ``SURROUNDED").

\section{Extra Visual Comparisons}
\begin{figure*}[h]
\centering
\includegraphics[width=1\textwidth]{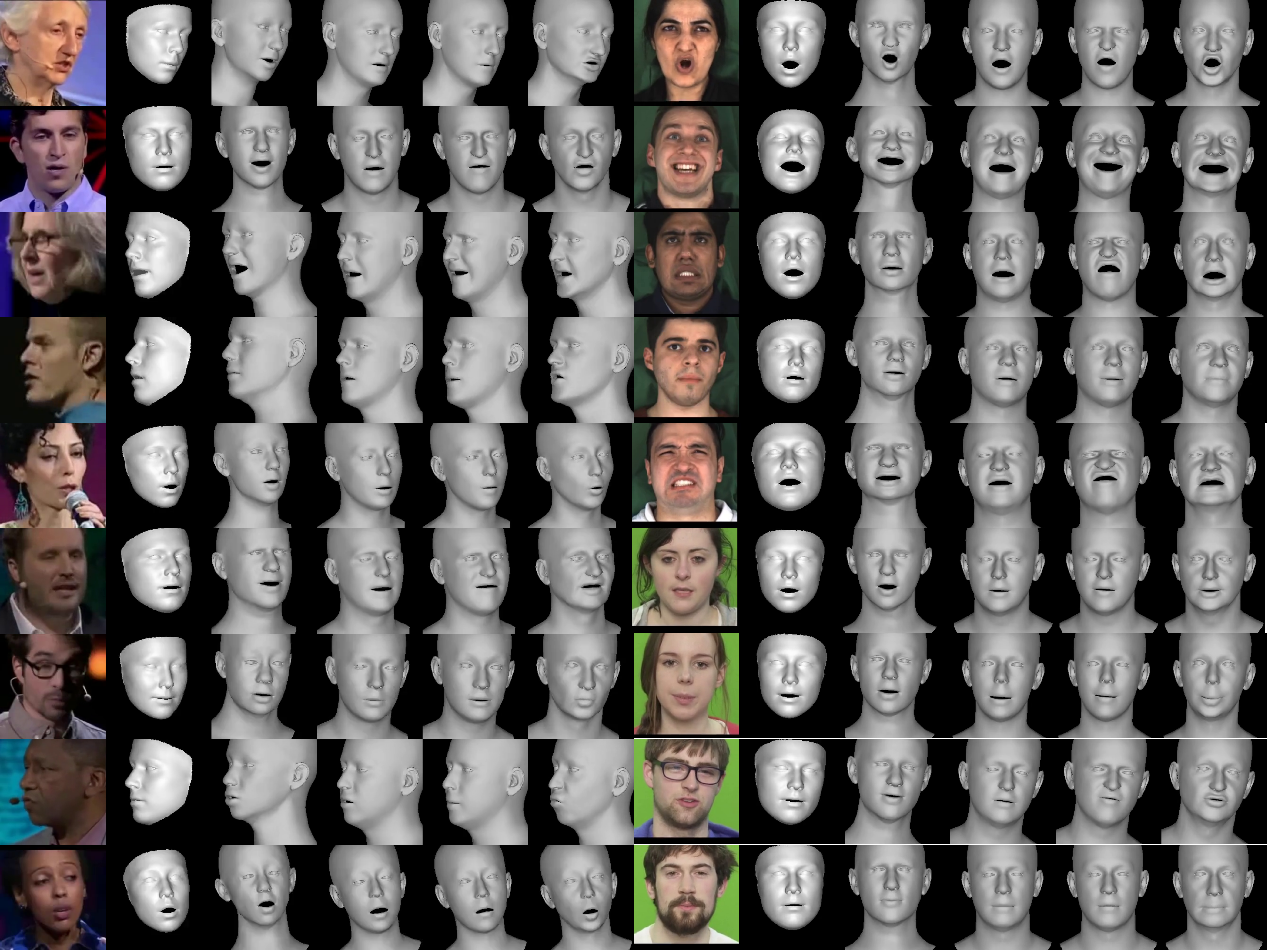}

\caption{More visual results and comparisons with other methods on the LRS3, MEAD, and TCDTIMIT datasets. From left to right: original footage, 3DDFAv2~\cite{3ddfa_cleardusk}, DAD~\cite{dad3dheads}, DECA~\cite{feng2021learning}, EMOCA~\cite{danvevcek2022emoca}, ours. In the accompanying video you will also find many video examples with sound.}
\label{fig:visual_results_more}
\end{figure*}

Finally, in Figure~\ref{fig:visual_results_more} we show more visual comparisons with 3DDFAv2~\cite{3ddfa_cleardusk}, DAD~\cite{dad3dheads}, DECA~\cite{feng2021learning}, and EMOCA~\cite{danvevcek2022emoca}. We also refer to the acoompanying video of the supplementary material where you will also find many video examples with sound.

\end{document}